\documentclass{article}
\usepackage{ijcai25}

\usepackage{times}
\usepackage{soul}
\usepackage{url}
\usepackage[hidelinks]{hyperref}
\usepackage[utf8]{inputenc}
\usepackage[small]{caption}
\usepackage{graphicx}
\usepackage{amsmath}
\usepackage{amsthm}
\usepackage{booktabs}
\usepackage{algorithm}
\usepackage{algorithmic}
\usepackage{multicol}
\usepackage{multirow}
\usepackage{amssymb}
\usepackage[switch]{lineno}
\usepackage[capitalize]{cleveref}
\Crefname{table}{Table}{Tables}
\crefname{table}{Table}{Tables}
\Crefname{figure}{Figure}{Figures}
\crefname{figure}{Figure}{Figures}
\usepackage{algorithm,algorithmic}

\title{SCOUT: Semi-supervised Camouflaged Object Detection by Utilizing Text and Adaptive Data Selection}

\author{
Weiqi Yan$^1$
\and
Lvhai Chen$^1$\and
Shengchuan Zhang$^{1}$\thanks{Corresponding author}\and
Yan Zhang$^1$\And
Liujuan Cao$^1$\\
\affiliations
$^1$Key Laboratory of Multimedia Trusted Perception and Efficient Computing,\\ Ministry of Education of China, Xiamen University, 361005, P.R. China. \\
\emails
weiqi\_yan@outlook.com,
lvhaichen2002@gmail.com,
zsc\_2016@xmu.edu.cn
}

\begin{document}

\maketitle

\begin{abstract}
    The difficulty of pixel-level annotation has significantly hindered the development of the Camouflaged Object Detection (COD) field.
    To save on annotation costs, previous works leverage the semi-supervised COD framework that relies on a small number of labeled data and a large volume of unlabeled data. 
    We argue that there is still significant room for improvement in the effective utilization of unlabeled data. To this end, we introduce a \textbf{S}emi-supervised \textbf{C}amouflaged \textbf{O}bject Detection by \textbf{U}tilizing \textbf{T}ext and Adaptive Data Selection (\textbf{SCOUT}).
    It includes an Adaptive Data Augment and Selection (ADAS) module and a Text Fusion Module (TFM). 
    The ADSA module selects valuable data for annotation through an adversarial augment and sampling strategy.
    The TFM module further leverages the selected valuable data by combining camouflage-related knowledge and text-visual interaction. 
    To adapt to this work, we build a new dataset, namely RefTextCOD.
    Extensive experiments show that the proposed method surpasses previous semi-supervised methods in the COD field and achieves state-of-the-art performance.
    Our code will be released at \url{https://github.com/Heartfirey/SCOUT}. 
\end{abstract}

\section{Introduction}

Camouflaged object detection (COD) \cite{Fan2020b,Fan2022} aims at segmenting objects that are visually concealed in their surroundings, which has important applications in several fields, such as military \cite{Cannaday2023}, environmental monitoring \cite{Yadav2018}, urban security \cite{Li2024} etc.
Existing COD methods always require large amounts of labeled data to segment camouflaged objects precisely \cite{Mei2021}. 

\begin{figure}
    \centering
    \includegraphics[width=\linewidth]{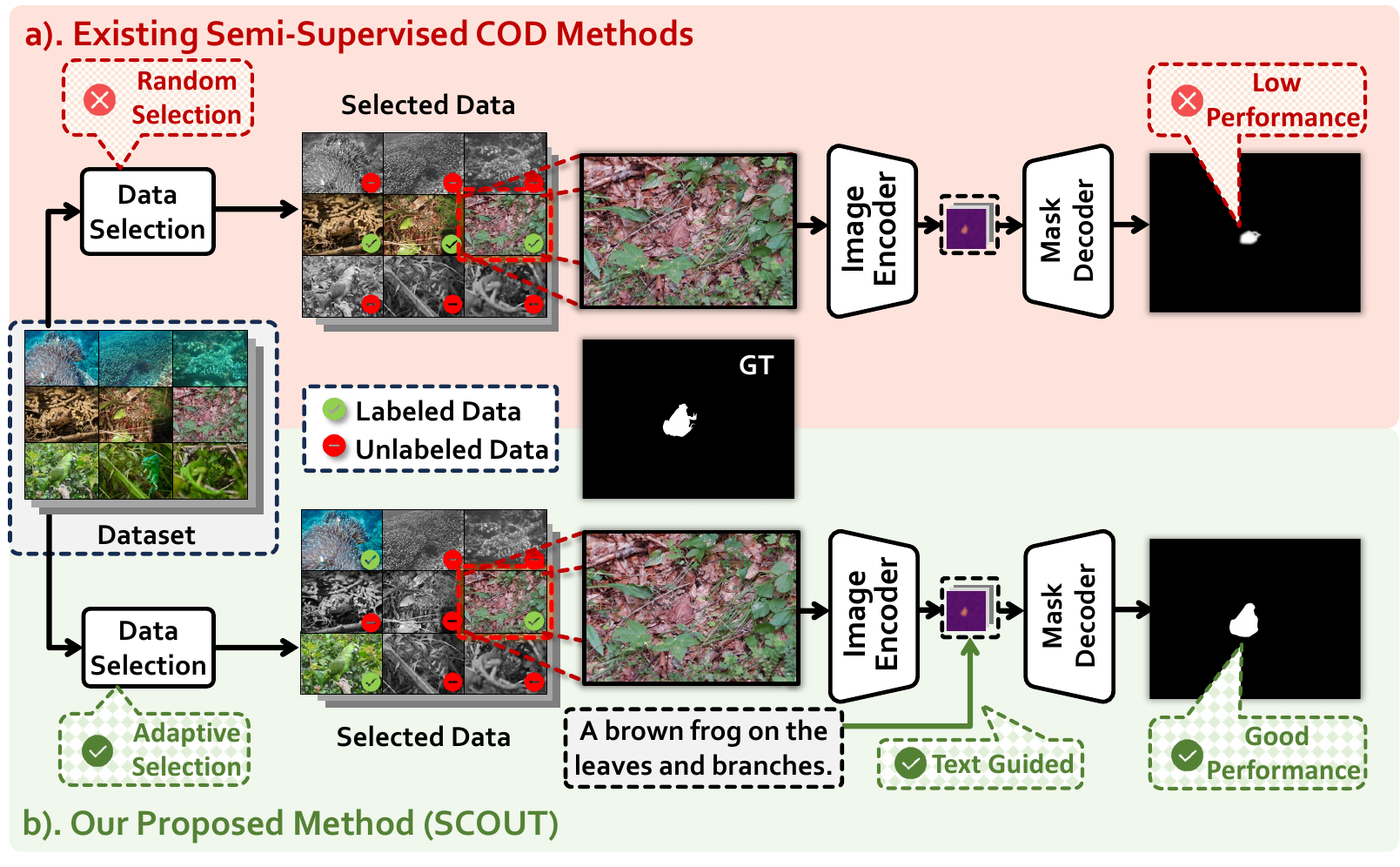}
    \caption{Comparison between the proposed method SCOUT and previous semi-supervised COD methods. The proposed SCOUT introduces an adaptive selection strategy and text guidance strategy and achieves a better segmentation performance. }
    \label{fig:cover}
\end{figure}

However, camouflaged objects often employ complex camouflage strategies to blend deeply with their background. Such camouflage strategies make pixel-level camouflaged object annotations difficult to obtain, and annotating the entire dataset requires a significantly greater cost. To reduce the annotation costs, semi-supervised COD methods utilize only a small amount of labeled data alongside a large amount of unlabelled data. 
However, as shown in \cref{fig:cover}, we identified the following two main issues with existing methods \cite{Lai2024,Zhang2025}: 1). Most existing methods rely on random sampling to select a portion of the data as the labeled set, without thoroughly considering the quality of the selected data. This often results in the annotation of some meaningless data. 2). For the selected small amount of labeled data, existing models struggle to fully learn camouflage-related knowledge, resulting in poor segmentation performance.

Therefore, we argue that there is still significant room for improvement in the effectiveness of unlabeled data utilization. As shown in \cref{fig:cover}, we propose a SCOUT model that introduces an adaptive selection strategy and a text guidance strategy. Specifically, the adaptive selection strategy is implemented by the Adaptive Data Augmentation and Selection(ADAS) module. The ADAS module adaptively selects valuable training data through the Adaptive Data Augmentation (ADA) and the Adaptive Data Selection (ADS) component. The text guidance strategy is implemented by the Text Fusion Module (TFM). To help the model fully learn camouflage-related knowledge, TFM introduces referring text assistance and enhances representation ability, thereby improving the model's performance across various camouflage scenarios. 

By annotating precise image-level referring text for existing mainstream COD datasets, we develop a new dataset called RefTextCOD. Our comprehensive evaluations on this dataset demonstrate substantial enhancements over existing semi-supervised COD models. Especially, our method improves the Mean Absolute Error (MAE) by $52.0\%$ and the S-Measure by $19.1\%$ compared to previous SOTA semi-supervised COD methods. Additionally, our approach outperforms some supervised COD methods, highlighting its greater practical applicability.

Our main contributions are summarized as follows:
\begin{itemize}
    \item We proposed an innovative semi-supervised COD model SCOUT, and extensive experiments have demonstrated its high performance and effectiveness.
    \item To avoid meaningless data annotations, we proposed an ADAS module, which selects valuable data through an adversarial augmentation and sampling strategy. 
    \item To fully utilize the referring text, we proposed a TFM module, which leverages the selected valuable data by combining camouflage-related knowledge and text-visual interaction. 
    \item We build a new dataset, namely RefTextCOD. We performed image-level referring text annotations on existing mainstream camouflaged object detection (COD) datasets, providing a data foundation for this paper and future explorations of other COD tasks.
\end{itemize}
\section{Related Works}

\subsection{Camouflaged Object Detection}
Camouflaged object detection (COD) has a long-standing history, and deep learning-based COD methods have seen rapid development in recent years. Existing fully-supervised COD methods have already achieved great performance by employing multiple strategies. For example, \cite{Pang2022,Pang2023} extract multi-scale features from the backbone and design strategies for fusion. \cite{Fan2022,Jia2022,Zhang2022} further use multi-stage refine. Some methods introduce additional information, \textit{e.g.} boundary guidance \cite{Sun2022,Ji2023,Zhai2021}, texture clues \cite{Ji2023,Zhu2021,Ren2023}, and other information such as frequency domain and depth \cite{Zhong2022,Lin2023}. 

The fully-supervised COD methods heavily rely on a large amount of labeled data, while pixel-level annotation incurs significant costs. As a result, some unsupervised, weakly-supervised, and semi-supervised methods have emerged. UCOS-DA \cite{Zhang2023a} is the first unsupervised COD method that addresses the task as a domain adaptation problem. SCOD \cite{Zhang2025} is the first weakly supervised COD method, which introduces a novel feature-guided loss and consistency loss with a new scribble learning approach. Semi-supervised COD methods have also received considerable attention from researchers. This paper focuses on investigating the rationality of data selection and the sufficiency of data utilization in semi-supervised COD methods.

\subsection{Semi-Supervised Camouflaged Object Detection}

In traditional fully-supervised learning, models require extensive labeled data for training to achieve optimal performance. However, obtaining labeled data in practical applications is often costly and time-consuming. Semi-supervised learning (SSL) enhances the model's generalization capability by combining labeled data and unlabeled data \cite{Chen2023,Grandvalet2004,Mi2022}, which effectively addresses the challenges of acquiring labeled data. Some previous works \cite{Chen2021,Sohn2020a,Wang2022a} introduce the pseudo-labeling mechanism, where a teacher model is trained using a small amount of labeled data, and then the teacher model is used to produce the pseudo labels of the unlabeled data for the subsequent training of the student model. Semi-supervised learning has shown significant potential in various fields. 

In semi-supervised COD methods, CamoTeacher \cite{Lai2024} applies different augmentation strategies to teacher-student networks and introduces a dual-rotation consistency loss. SCOD-ND \cite{Fu2024} proposes a window-based voting strategy and an ensemble learning algorithm to eliminate noise from labels. However, these methods typically split the dataset into labeled sets and unlabeled sets by random sampling, without considering the value of the selected data. Moreover, they do not make full use of the labeled data. In this paper, We adaptively select valuable samples for training and introduce referring text to assist the model in learning camouflage-related knowledge from the labeled data.

\subsection{Referring Camouflaged Object Detection}

The concept of Referring Camouflaged Object Detection (Ref-COD) was first proposed by \cite{Zhang2023}, which leverages a batch of images as the referring information to guide the identification of the specified camouflaged objects. With the development of MLLMs, the rich intrinsic knowledge that MLLMs learned from massive amounts of data can be used to augment a variety of downstream tasks. Recently works \cite{Cheng2023,Hu2024} have extended this concept by utilizing MLLMs and designing a series of prompts to assist the COD task. 
In this paper, we focus on how to use precise referring text to assist the model in learning camouflage-related knowledge from labeled data.

\subsection{Active Learning}

Active learning in the context of deep learning is a strategy designed to reduce the labeling cost by allowing the model to selectively query the most informative samples from an unlabeled dataset for annotation. It is particularly useful in scenarios where data labeling is expensive, time-consuming, or requires expert knowledge. Active learning approaches can be divided into membership query synthesis \cite{Angluin1988,King2004}, stream-based selective sampling \cite{ArgamonEngelson1999}, and pool-based active learning from application scenarios \cite{Settles2009}. 
In this paper, we use an adversarial augmentation and selection strategy to select valuable data that the model can partially understand but still needs further learning.
\begin{figure*}
    \centering
    \includegraphics[width=\linewidth]{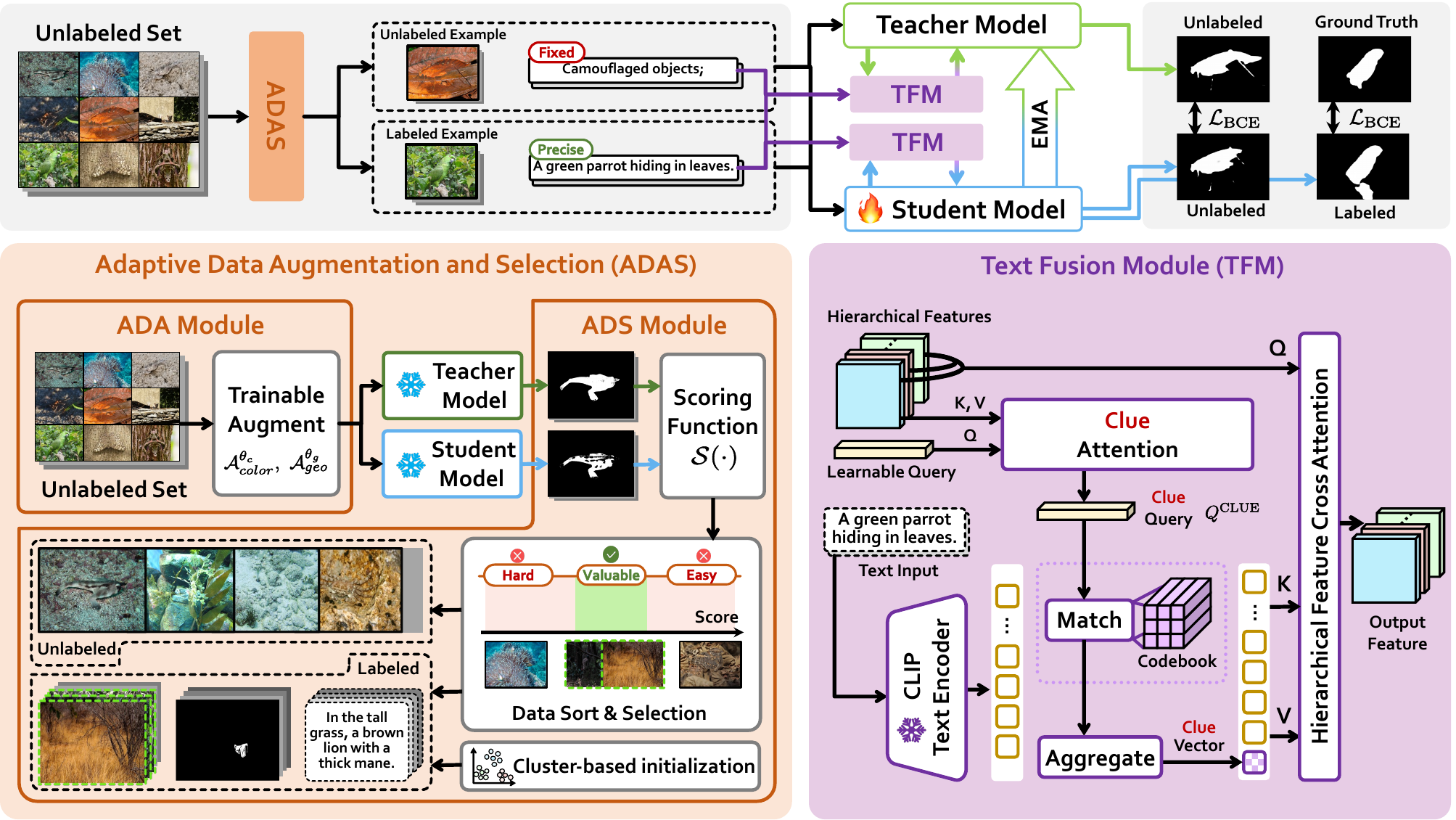}
    \caption{\textbf{Overview of the proposed SCOUT.} The SCOUT is an innovative semi-supervised COD model, which consists of an ADAS module and a TFM module. }
    \label{fig:main_structure}
\end{figure*}

\section{RefTextCOD Dataset}

To adapt to this work, we build a new dataset, namely RefTextCOD. The motivation behind this proposed dataset is that existing datasets do not have image-level referring text. 

To achieve a fair and effective comparison with existing methods, we expect to be able to construct text-based referring COD experiments in settings that are as similar as possible. Referring to \cite{Chen2022,Fan2022}, we used the mainstream COD datasets: CHAMELEON (\cite{Wu2019}), CAMO (\cite{Yan2021,Le2019}), COD10K (\cite{Fan2020b}), NC4K (\cite{Le2019}) as the base image data. This allows us to focus more on annotations without collecting camouflaged images from scratch.

It is time-consuming to annotate camouflaged objects with captions, and manual annotating tends to lead to inconsistent quality. To ensure that the camouflaged object captions contain meaningful information for COD tasks (\textit{i.e.} texture, color, and shape), we used two vision language models (\textit{e.g.} QwenVL \cite{Bai2023} and GPT4-Vision \cite{OpenAI2024}) to generate a summary first. Specifically, we design a series of prompts with contextual logic: 1). First, the VLM model will be guided to locate the object and justify its classes; 2). Then, the model is directed to characterize the physical properties of foreground objects and background. 3). Finally, we request the model to aggregate and streamline all the features to generate complete referring text. 

Finally, we employing manual screening and conducted a thorough review and revision of all annotations to ensure their accuracy and quality. As a result, we annotated \textbf{9,487} images from four datasets and organized them into a new dataset called RefTextCOD. The prompt used in the above process, the whole pipeline will be provided in the appendix.

\section{Methods}


The overview of SCOUT is depicted in \cref{fig:main_structure}. In \cref{sec:ADAS}, we introduce the Adaptive Data Augmentation and Selection (ADAS) module. The module selects valuable data for annotation through an Adaptive Data Augment (ADA) component and an Adaptive Data Selection (ADS) component. In \cref{sec:TFM}, we present the Text Fusion Module (TFM). Finally, we provide the details of the total loss and training process in \cref{sec:totloss}.




\subsection{Adaptive Data Augmentation and Selection}
\label{sec:ADAS}
During this phase, our objective is to select valuable data from an unlabeled set $\mathcal{D}_{\mathrm{U}} = \{\mathrm{X}_{u,j}^I\}_{j=1}^{M}$ for semi-supervised training, represented by the ADAS module.
The ADAS module contains two sub-components: an Adaptive Data Augmentation (ADA) component and an Adaptive Data Selection (ADS) component. The ADA is a trainable data augmentation module through an adversarial augment strategy, while the ADS finds valuable data via an adversarial sampling strategy.

Formally, given the unlabeled set $\mathcal{D}_{\mathrm{U}} = \{\mathrm{X}_{u,j}^I\}_{j=1}^{M}$, the ADA component adopts a set of parameterizable data augmenter \cite{Suzuki2022} $\{\mathcal{A}^{\theta_c}_{color}, \mathcal{A}^{\theta_g}_{geo}\}$ (\textit{i.e.} , color augmentation and geometric transformation) to augment each unlabeled data $\mathrm{X}^I_{u,j}$,
\begin{equation}
    \mathrm{X}_{u,j}^{I, \mathrm{Aug}} = \mathcal{A}^{\theta_c}_{color} \circ \mathcal{A}^{\theta_g}_{geo} (\mathrm{X}^I_{u,j}),
    \label{eq:augmenter}
\end{equation}
where $\mathrm{X}_{u,j}^{I, \mathrm{Aug}}$ denotes the augmented unlabeled data. We feed $\mathrm{X}_{u,j}^{I, \mathrm{Aug}}$ into both teacher and student models, parameterized by $\mathrm{\theta}_T$ and $\mathrm{\theta}_S$, to obtain segmentation masks $\mathrm{\hat{Y}}_{u, j}^T$ and $\mathrm{\hat{Y}}_{u, j}^S$, 
\begin{equation}
    \mathrm{\hat{Y}}_{u, j}^T = \mathcal{F}(\mathrm{X}_{u,j}^{I,\mathrm{Aug}}; \mathrm{\theta}_{T}),\ \mathrm{\hat{Y}}_{u, j}^S = \mathcal{F}(\mathrm{X}_{u,j}^{I,\mathrm{Aug}}; \mathrm{\theta}_{S}),
    \label{eq:score calculation}
\end{equation}
where $\mathcal{F}(\mathrm{X}_{u,j}^{I,\mathrm{Aug}}; \mathrm{\theta}_{T}), \mathcal{F}(\mathrm{X}_{u,j}^{\mathrm{I,Aug}}; \mathrm{\theta}_{S})$ denote the predictions of the teacher model and student model, respectively, on augmented unlabeled data $\mathrm{X}_{u,j}^{I,\mathrm{Aug}}$. 

After obtaining the segmentation masks $\mathrm{\hat{Y}}_{u, j}^T$ and $\mathrm{\hat{Y}}_{u, j}^S$, the ADS calculates the score of each unlabeled data $\mathrm{X}_{u,j}^{I}$ and selects valuable data for annotation. Specifically, we first use a scoring function $\mathcal{S}(\cdot)$ to calculate the unlabeled data $\mathrm{X}^I_{u,j}$ scores,
\begin{equation}
    \mathcal{S}(\mathrm{X}_{u,j}^I) = \mathrm{SSIM}(\mathrm{\hat{Y}}_{u, j}^T, \mathrm{\hat{Y}}_{u, j}^S) - \mathrm{MAE}(\mathrm{\hat{Y}}_{u, j}^T, \mathrm{\hat{Y}}_{u, j}^S),
    \label{eq:ads_scores}
\end{equation}
where $\mathrm{SSIM}(\cdot)$ denotes the structural similarity metrics, and $\mathrm{MAE}(\cdot)$ denotes the mean absolute error. The ADS component uses Kernel Density Estimation (KDE) normalization to normalize all scores and uniformly map them to the $[0, 1]$ interval, then it sorts the unlabeled set $\mathcal{D}_{\mathrm{U}}$ based on these scores. At this point, data with scores close to 0 are considered hard data, while those close to 1 are considered easy data. Among these ordered data, extremely low-scoring hard data are too complex for the model or contain noise that hinders its convergence. On the other hand, extremely high-scoring easy data are of no value to train the model and contribute to overfitting. Therefore, the ADS component selects data with moderate scores that are close to 0.5, which the model can partially understand but still needs further learning. Specifically, the ADS component samples the unlabeled data in unlabeled set  $\mathcal{D}_{\mathrm{U}} = \{\mathrm{X}_{u,j}^I\}_{j=1}^{M}$ with scores close to 0.5.  Those selected unlabeled data $\mathrm{X}_{l,i}^I$ will undergo referring text annotation $\mathrm{X}_{l,i}^T$ and segmentation mask annotation $\mathrm{Y}_{l,i}$ to construct the labeled set $\mathcal{D}_{\mathrm{L}} = \{\mathrm{X}_{l,i}^I, \mathrm{X}_{l,i}^T, \mathrm{Y}_{l,i}\}_{i=1}^{N}$ for next training phase. Those remaining unlabeled data will form a new unlabeled set $\mathcal{D}_{\mathrm{R}} = \{\mathrm{X}_{u,o}^I\}_{o=1}^{O}$.

To obtain a pre-trained teacher and student model, we initialize the labeled set $\mathcal{D}_{\mathrm{L}}$ by considering the diversity of data in $\mathcal{D}_{\mathrm{U}}$. Specifically, we perform K-Means clustering using the color, texture, and frequency domain features of data in $\mathcal{D}_{\mathrm{U}}$. Then we select the data near the clustering centers to construct the initial labeled set $\mathcal{D}_{\mathrm{L}}$.

The ADA component adopts an adversarial manner for training. Specifically, given the labeled set $\mathcal{D}_{\mathrm{L}}$, we use the pre-trained augmenter $\{\mathcal{A}^{\theta_c}_{color}, \mathcal{A}^{\theta_g}_{geo}\}$ to obtain the augmented input $\mathrm{X}_{l,i}^{I,\mathrm{Aug}}$ by \cref{eq:augmenter}. Then we train the teacher and student models with Binary Cross-Entropy (BCE) loss $\mathcal{L}_{\mathrm{BCE}}$. Following \cite{Suzuki2022}, the training objective of augmenter loss $\mathcal{L}_{\mathrm{Aug}}$ is to minimize the loss of the teacher model and maximize the loss of the student model,
\begin{equation}
\begin{aligned}
    \mathcal{L}_{\mathrm{Aug}} = &\mathcal{L}_{\mathrm{BCE}}(\mathcal{F}(\mathrm{X}_{l,i}^{I,\mathrm{Aug}}; \mathrm{\theta}_{T}), \mathrm{Y}_{l,i}) \\
    &- \mathcal{L}_{\mathrm{BCE}}(\mathcal{F}(\mathrm{X}_{l,i}^{I,\mathrm{Aug}}; \mathrm{\theta}_{S}), \mathrm{Y}_{l,i}).
\end{aligned}
\end{equation}

\def\metricsCOD{
    &$\mathcal{S}_m\uparrow$
    &$\mathcal{F}_{\beta}^{\omega}\uparrow$
    &$\mathcal{F}_{\beta}^{m}\uparrow$
    &$\mathcal{E}_{\phi}^{m}\uparrow$
    &$\mathcal{E}_{\phi}^{x}\uparrow$
    &$\mathcal{M}\downarrow$
}

\def\metricsCODrline{
    & \multicolumn{1}{l}{$\mathcal{S}_m\uparrow$}
    & \multicolumn{1}{l}{$\mathcal{F}_{\beta}^{\omega}\uparrow$}
    & \multicolumn{1}{l}{$\mathcal{F}_{\beta}^{m}\uparrow$}
    & \multicolumn{1}{l}{$\mathcal{E}_{\phi}^{m}\uparrow$}
    & \multicolumn{1}{l}{$\mathcal{E}_{\phi}^{x}\uparrow$}
    & \multicolumn{1}{l|}{$\mathcal{M}\downarrow$}
}

\def\metricsCODnline{
    & \multicolumn{1}{l}{$\mathcal{S}_m\uparrow$}
    & \multicolumn{1}{l}{$\mathcal{F}_{\beta}^{\omega}\uparrow$}
    & \multicolumn{1}{l}{$\mathcal{F}_{\beta}^{m}\uparrow$}
    & \multicolumn{1}{l}{$\mathcal{E}_{\phi}^{m}\uparrow$}
    & \multicolumn{1}{l}{$\mathcal{E}_{\phi}^{x}\uparrow$}
    & \multicolumn{1}{l}{$\mathcal{M}\downarrow$}
}
\begin{table*}[t!]
\setlength{\belowcaptionskip}{0cm}   
\renewcommand{\arraystretch}{1.1}
\renewcommand{\tabcolsep}{3pt}
\centering
\resizebox{\linewidth}{!}{
\begin{tabular}{ccccccccccccccccccc}
\toprule
\multicolumn{19}{c}{\textbf{CHAMELEON (76)}} \\ \hline
\multicolumn{1}{c|}{\multirow{2}{*}{\textbf{Methods}}} & \multicolumn{6}{c|}{\textbf{1\% (41)}} & \multicolumn{6}{c|}{\textbf{5\% (202)}} & \multicolumn{6}{c}{\textbf{10\% (404)}} \\ \cline{2-19} 
\multicolumn{1}{c|}{} & \multicolumn{1}{c}{$\mathcal{S}_m\uparrow$} & \multicolumn{1}{c}{$\mathcal{F}_{\beta}^{\omega}\uparrow$} & \multicolumn{1}{c}{$\mathcal{F}_{\beta}^{m}\uparrow$} & \multicolumn{1}{c}{$\mathcal{E}_{\phi}^{m}\uparrow$} & \multicolumn{1}{c}{$\mathcal{E}_{\phi}^{x}\uparrow$} & \multicolumn{1}{l|}{$\mathcal{M}\downarrow$} & \multicolumn{1}{c}{$\mathcal{S}_m\uparrow$} & \multicolumn{1}{c}{$\mathcal{F}_{\beta}^{\omega}\uparrow$} & \multicolumn{1}{c}{$\mathcal{F}_{\beta}^{m}\uparrow$} & \multicolumn{1}{c}{$\mathcal{E}_{\phi}^{m}\uparrow$} & \multicolumn{1}{c}{$\mathcal{E}_{\phi}^{x}\uparrow$} & \multicolumn{1}{l|}{$\mathcal{M}\downarrow$} & \multicolumn{1}{c}{$\mathcal{S}_m\uparrow$} & \multicolumn{1}{c}{$\mathcal{F}_{\beta}^{\omega}\uparrow$} & \multicolumn{1}{c}{$\mathcal{F}_{\beta}^{m}\uparrow$} & \multicolumn{1}{c}{$\mathcal{E}_{\phi}^{m}\uparrow$} & \multicolumn{1}{c}{$\mathcal{E}_{\phi}^{x}\uparrow$} & \multicolumn{1}{c}{$\mathcal{M}\downarrow$}  \\ \hline
\multicolumn{1}{c|}{Mean Teacher \cite{Lai2024}} & .537 & .199 & .229 & .418 & .636 & \multicolumn{1}{c|}{.204} & .611 & .309 & .353 & .524 & .745 & \multicolumn{1}{c|}{.137} & .679 & .450 & .512 & .650 & .812 & .102 \\
\multicolumn{1}{c|}{CamoTeacher \cite{Lai2024}} & .652 & .472 & .558 & .714 & .762 & \multicolumn{1}{c|}{.093} & .729 & .587 & .656 & .785 & .822 & \multicolumn{1}{c|}{.070} & .756 & .617 & .684 & .813 & .851 & .065 \\
\multicolumn{1}{c|}{SCOD-ND \cite{Fu2024}} & - & - & - & - & - & \multicolumn{1}{c|}{-} & - & - & - & - & - & \multicolumn{1}{c|}{-} & .850 & .773 & - & \textbf{.928} & - & .036 \\
\multicolumn{1}{c|}{\textbf{SCOUT} (Ours) \dag} & \underline{.846}  &  \underline{.773}  & \underline{.806} & \underline{.885} & \underline{.896} &\multicolumn{1}{c|}{ \underline{.039} }& \underline{.877}  &  \underline{.821}  & \underline{.850} & \underline{.929} & \underline{.941} &\multicolumn{1}{c|}{ \underline{.028} }& \underline{.874}  &  \underline{.815}  & \underline{.838} & .916 & \underline{.925} & \underline{.027} \\
\multicolumn{1}{c|}{\textbf{SCOUT} (Ours) \ddag} & \textbf{.847}  &  \textbf{.777}  & \textbf{.810} & \textbf{.887} & \textbf{.898} & \multicolumn{1}{c|}{\textbf{.038}} &  \textbf{.880}  &  \textbf{.827}  & \textbf{.852} & \textbf{.935} & \textbf{.944} & \multicolumn{1}{c|}{\textbf{.027}} & \textbf{.876}  &  \textbf{.817}  & \textbf{.837} & \underline{.922} & \textbf{.932} & \textbf{.026} \\
\midrule
\toprule
\multicolumn{19}{c}{\textbf{CAMO (250)}} \\ \hline
\multicolumn{1}{c|}{\multirow{2}{*}{\textbf{Methods}}} & \multicolumn{6}{c|}{\textbf{1\% (41)}} & \multicolumn{6}{c|}{\textbf{5\% (202)}} & \multicolumn{6}{c}{\textbf{10\% (404)}} \\ \cline{2-19} 
\multicolumn{1}{c|}{} & \multicolumn{1}{c}{$\mathcal{S}_m\uparrow$} & \multicolumn{1}{c}{$\mathcal{F}_{\beta}^{\omega}\uparrow$} & \multicolumn{1}{c}{$\mathcal{F}_{\beta}^{m}\uparrow$} & \multicolumn{1}{c}{$\mathcal{E}_{\phi}^{m}\uparrow$} & \multicolumn{1}{c}{$\mathcal{E}_{\phi}^{x}\uparrow$} & \multicolumn{1}{l|}{$\mathcal{M}\downarrow$} & \multicolumn{1}{c}{$\mathcal{S}_m\uparrow$} & \multicolumn{1}{c}{$\mathcal{F}_{\beta}^{\omega}\uparrow$} & \multicolumn{1}{c}{$\mathcal{F}_{\beta}^{m}\uparrow$} & \multicolumn{1}{c}{$\mathcal{E}_{\phi}^{m}\uparrow$} & \multicolumn{1}{c}{$\mathcal{E}_{\phi}^{x}\uparrow$} & \multicolumn{1}{l|}{$\mathcal{M}\downarrow$} & \multicolumn{1}{c}{$\mathcal{S}_m\uparrow$} & \multicolumn{1}{c}{$\mathcal{F}_{\beta}^{\omega}\uparrow$} & \multicolumn{1}{c}{$\mathcal{F}_{\beta}^{m}\uparrow$} & \multicolumn{1}{c}{$\mathcal{E}_{\phi}^{m}\uparrow$} & \multicolumn{1}{c}{$\mathcal{E}_{\phi}^{x}\uparrow$} & \multicolumn{1}{c}{$\mathcal{M}\downarrow$}  \\ \hline
\multicolumn{1}{c|}{Mean Teacher \cite{Lai2024}} & .518 & .207 & .227 & .399 & .620 & \multicolumn{1}{c|}{.226} & .575 & .286 & .322 & .482 & .708 & \multicolumn{1}{c|}{.184} & .625 & .397 & .454 & .578 & .773 & .150 \\
\multicolumn{1}{c|}{CamoTeacher \cite{Lai2024}}  & .621 & .456 & .545 & .669 & .736 & \multicolumn{1}{c|}{.136} & .669 & .523 & .601 & .711 & .775 & \multicolumn{1}{c|}{.122} & .701 & .560 & .635 & .742 & .795 & .112 \\
\multicolumn{1}{c|}{SCOD-ND \cite{Fu2024}} & - & - & - & - & - & \multicolumn{1}{c|}{-} & - & - & - & - & - & \multicolumn{1}{c|}{-} & .789 & .732 & - & .859 & - & .077 \\
\multicolumn{1}{c|}{\textbf{SCOUT} (Ours) \dag}  & \textbf{.798}  &  \underline{.732}  & \textbf{.782} & \underline{.845} & \underline{.864} & \multicolumn{1}{c|}{\textbf{.076}} & \underline{.847}  &  \underline{.802}  & \underline{.839} & \underline{.897} & \textbf{.909} & \multicolumn{1}{c|}{\underline{.055}} & \underline{.847}  &  \underline{.812}  & \underline{.849} & \underline{.901} & \underline{.912} & \underline{.052} \\
\multicolumn{1}{c|}{\textbf{SCOUT} (Ours) \ddag} & \underline{.795}  &  \textbf{.732}  & \underline{.780} & \textbf{.845} & \textbf{.864} & \multicolumn{1}{c|}{\underline{.077}} & \textbf{.848}  &  \textbf{.807}  & \textbf{.841} & \textbf{.901} & \underline{.908} & \multicolumn{1}{c|}{\textbf{.054}} & \textbf{.859}  &  \textbf{.828}  & \textbf{.857} & \textbf{.919} & \textbf{.925} & \textbf{.047} \\
\midrule
\toprule
\multicolumn{19}{c}{\textbf{COD10K (2026)}} \\ \hline
\multicolumn{1}{c|}{\multirow{2}{*}{\textbf{Methods}}} & \multicolumn{6}{c|}{\textbf{1\% (41)}} & \multicolumn{6}{c|}{\textbf{5\% (202)}} & \multicolumn{6}{c}{\textbf{10\% (404)}} \\ \cline{2-19} 
\multicolumn{1}{c|}{} & \multicolumn{1}{c}{$\mathcal{S}_m\uparrow$} & \multicolumn{1}{c}{$\mathcal{F}_{\beta}^{\omega}\uparrow$} & \multicolumn{1}{c}{$\mathcal{F}_{\beta}^{m}\uparrow$} & \multicolumn{1}{c}{$\mathcal{E}_{\phi}^{m}\uparrow$} & \multicolumn{1}{c}{$\mathcal{E}_{\phi}^{x}\uparrow$} & \multicolumn{1}{l|}{$\mathcal{M}\downarrow$} & \multicolumn{1}{c}{$\mathcal{S}_m\uparrow$} & \multicolumn{1}{c}{$\mathcal{F}_{\beta}^{\omega}\uparrow$} & \multicolumn{1}{c}{$\mathcal{F}_{\beta}^{m}\uparrow$} & \multicolumn{1}{c}{$\mathcal{E}_{\phi}^{m}\uparrow$} & \multicolumn{1}{c}{$\mathcal{E}_{\phi}^{x}\uparrow$} & \multicolumn{1}{l|}{$\mathcal{M}\downarrow$} & \multicolumn{1}{c}{$\mathcal{S}_m\uparrow$} & \multicolumn{1}{c}{$\mathcal{F}_{\beta}^{\omega}\uparrow$} & \multicolumn{1}{c}{$\mathcal{F}_{\beta}^{m}\uparrow$} & \multicolumn{1}{c}{$\mathcal{E}_{\phi}^{m}\uparrow$} & \multicolumn{1}{c}{$\mathcal{E}_{\phi}^{x}\uparrow$} & \multicolumn{1}{c}{$\mathcal{M}\downarrow$}  \\ \hline
\multicolumn{1}{c|}{Mean Teacher \cite{Lai2024}} & .546 & .168 & .226 & .441 & .633 & \multicolumn{1}{c|}{.161} & .621 & .272 & .343 & .555 & .732 & \multicolumn{1}{c|}{.107} & .683 & .404 & .482 & .666 & .799 & .078 \\
\multicolumn{1}{c|}{CamoTeacher \cite{Lai2024}} & .699 & .517 & .582 & .788 & .797 & \multicolumn{1}{c|}{.062} & .745 & .583 & .644 & .827 & .840 & \multicolumn{1}{c|}{.050} & .759 & .594 & .652 & .836 & .854 & .049 \\
\multicolumn{1}{c|}{SCOD-ND \cite{Fu2024}} & - & - & - & - & - & \multicolumn{1}{c|}{-} & - & - & - & - & - & \multicolumn{1}{c|}{-} & .819 & .725 & - & .891 & - & {.033} \\
\multicolumn{1}{c|}{\textbf{SCOUT} (Ours) \dag} & \underline{.833}  &  \underline{.733}  & \underline{.770} & \underline{.891} & \underline{.899} & \multicolumn{1}{c|}{\underline{.031}} & \underline{.855}  &  \underline{.768}  & \underline{.801} & \underline{.913} & \underline{.924} & \multicolumn{1}{c|}{\underline{.027}} & \underline{.858}  &  \underline{.783}  & \textbf{.815} & \underline{.917} & \underline{.928} & \underline{.024} \\
\multicolumn{1}{c|}{\textbf{SCOUT} (Ours) \ddag} & \textbf{.834}  &  \textbf{.736}  & \textbf{.774} & \textbf{.892} & \textbf{.901} & \multicolumn{1}{c|}{\textbf{.031}} & \textbf{.859}  &  \textbf{.776}  & \textbf{.804} & \textbf{.919} & \textbf{.926} & \multicolumn{1}{c|}{\textbf{.026}} & \textbf{.861}  &  \textbf{.785}  & \underline{.811} & \textbf{.925} & \textbf{.932} & \textbf{.024} \\
\midrule
\toprule
\multicolumn{19}{c}{\textbf{NC4K (4121)}} \\ \hline
\multicolumn{1}{c|}{\multirow{2}{*}{\textbf{Methods}}} & \multicolumn{6}{c|}{\textbf{1\% (41)}} & \multicolumn{6}{c|}{\textbf{5\% (202)}} & \multicolumn{6}{c}{\textbf{10\% (404)}} \\ \cline{2-19} 
\multicolumn{1}{c|}{} & \multicolumn{1}{c}{$\mathcal{S}_m\uparrow$} & \multicolumn{1}{c}{$\mathcal{F}_{\beta}^{\omega}\uparrow$} & \multicolumn{1}{c}{$\mathcal{F}_{\beta}^{m}\uparrow$} & \multicolumn{1}{c}{$\mathcal{E}_{\phi}^{m}\uparrow$} & \multicolumn{1}{c}{$\mathcal{E}_{\phi}^{x}\uparrow$} & \multicolumn{1}{l|}{$\mathcal{M}\downarrow$} & \multicolumn{1}{c}{$\mathcal{S}_m\uparrow$} & \multicolumn{1}{c}{$\mathcal{F}_{\beta}^{\omega}\uparrow$} & \multicolumn{1}{c}{$\mathcal{F}_{\beta}^{m}\uparrow$} & \multicolumn{1}{c}{$\mathcal{E}_{\phi}^{m}\uparrow$} & \multicolumn{1}{c}{$\mathcal{E}_{\phi}^{x}\uparrow$} & \multicolumn{1}{l|}{$\mathcal{M}\downarrow$} & \multicolumn{1}{c}{$\mathcal{S}_m\uparrow$} & \multicolumn{1}{c}{$\mathcal{F}_{\beta}^{\omega}\uparrow$} & \multicolumn{1}{c}{$\mathcal{F}_{\beta}^{m}\uparrow$} & \multicolumn{1}{c}{$\mathcal{E}_{\phi}^{m}\uparrow$} & \multicolumn{1}{c}{$\mathcal{E}_{\phi}^{x}\uparrow$} & \multicolumn{1}{c}{$\mathcal{M}\downarrow$}  \\ \hline
\multicolumn{1}{c|}{Mean Teacher \cite{Lai2024}} & .541 & .213 & .258 & .424 & .637 & \multicolumn{1}{c|}{.193} & .634 & .355 & .420 & .556 & .767 & \multicolumn{1}{c|}{.140} & .700 & .492 & .565 & .670 & .827 & .109 \\
\multicolumn{1}{c|}{CamoTeacher \cite{Lai2024}} & .718 & .599 & .675 & .779 & .814 & \multicolumn{1}{c|}{.090} & .777 & .677 & .739 & .834 & .859 & \multicolumn{1}{c|}{.071} & .791 & .687 & .746 & .842 & .868 & .068 \\
\multicolumn{1}{c|}{SCOD-ND \cite{Fu2024}} & - & - & - & - & - & \multicolumn{1}{c|}{-} & - & - & - & - & - & \multicolumn{1}{c|}{-} & .838 & .787 & - & .903 & - & .046 \\
\multicolumn{1}{c|}{\textbf{SCOUT} (Ours) \dag} &  \underline{.849}  &  \underline{.791}  & \underline{.829} & \underline{.893} & \underline{.904} & \multicolumn{1}{c|}{\underline{.045}} & \underline{.874}  &  \underline{.826}  & \underline{.856} & \underline{.919} & \underline{.930} & \multicolumn{1}{c|}{\underline{.036}} & \underline{.873}  &  \underline{.833}  & \underline{.864} & \underline{.919} & \underline{.930} & \underline{.035} \\
\multicolumn{1}{c|}{\textbf{SCOUT} (Ours) \ddag} &  \textbf{.850}  &  \textbf{.793}  & \textbf{.832} & \textbf{.894} & \textbf{.905} & \multicolumn{1}{c|}{\textbf{.045}} & \textbf{.877}  &  \textbf{.832}  & \textbf{.858} & \textbf{.924} & \textbf{.931} & \multicolumn{1}{c|}{\textbf{.035}} & \textbf{.879}  &  \textbf{.839}  & \textbf{.864} & \textbf{.928} & \textbf{.935} & \textbf{.033} \\
\midrule
\end{tabular}}

\caption{\textbf{Quantitative comparison with existing methods on four COD benchmark testing sets including CHAMELEON, CAMO, COD10K and NC4K.} We provide experimental results under two test settings: \dag\ denotes that all data use fixed referring text during test time(\textit{i.e.} ``\textit{camouflaged objects; concealed objects; hidden objects;}"), and \ddag\ denotes that all data used precise referring text during test time. \textbf{Bold} indicates the best result in group settings, and \underline{underline} indicates the second-best result.}\label{tab:sota_cod}
\label{tab:main_results}
\end{table*}

\subsection{Text Fusion Module}
\label{sec:TFM}

The TFM module utilizes the precise referring text $\mathrm{X}_{l,i}^T$ in the labeled set $\mathcal{D}_{\mathrm{L}} = \{\mathrm{X}_{l,i}^I, \mathrm{X}_{l,i}^T, \mathrm{Y}_{l,i}\}_{i=1}^{N}$ to further extract the knowledge of camouflage object. Specifically, for input labeled data $\mathrm{X}_{l,i}^I$, with corresponding referring text annotations $\mathrm{X}_{l,i}^T$, the TFM utilizes CLIP-Text Encoder \cite{Radford2021} $\mathbf{CLIP}_{\mathcal{T}}(\cdot)$ to obtain text feature $\mathrm{F}_{l,i}^{T}$,
\begin{equation}
    \begin{aligned}
        \mathrm{F}_{l,i}^{T} = \mathbf{CLIP}_{\mathcal{T}}{(\mathrm{X}_{l,i}^T)}.
    \end{aligned}
\end{equation}

For input labeled data $\mathrm{X}_{l,i}^I$, assuming that the hierarchical features obtained by the Image Encoder are $\{\mathrm{F}_{l, i, k}^{I}\}, k \in [1, N_{E}]$,where $N_{E}$ is the number of hierarchical feature levels, we use a clue attention mechanism to generate the clue query $Q^{\mathrm{CLUE}}$, which highlights the camouflage-related region by the guidance of attention score $\mathrm{A}^{\mathrm{CAMO}}_{l, i}$.
\begin{equation}
    \begin{gathered}
    Q_C = q^{\mathrm{CAMO}}W_Q,\ K_C = \mathrm{F}_{l, i, N_{E}}^{I}W_K,\ V_C = \mathrm{F}_{l, i, N_{E}}^{I}W_V \\
    \mathrm{A}^{\mathrm{CAMO}}_{l, i} = \mathrm{Softmax}\left(\frac{Q_CK_C^{\top}}{\sqrt{d_h}}\right) \\
    Q^{\mathrm{CLUE}} = \mathrm{A}^{\mathrm{CAMO}}_{l, i}V_C,
    \end{gathered}
\end{equation}
where $q^{\mathrm{CAMO}}$ is a learnable query, and $W_Q, W_K, W_V$ denote the linear projection weight, respectively. $d_h$ represents the attention head dimension. In order to enable the clue attention mechanism to learn how to extract camouflage-related region features, the following loss function is leveraged,
\begin{equation}
    \mathcal{L}_{\mathrm{TFM}}^{\mathrm{Attn}} = \mathcal{L}_{\mathrm{BCE}}(\mathrm{A}^{\mathrm{CAMO}}_{l, i}, \mathrm{down}(\mathrm{Y}_{l,i})),
\end{equation}
where $\mathrm{down}(\cdot)$ denotes a downsampling function. Then we apply $Q^{\mathrm{CLUE}}$ to calculate the similarity scores $W^\mathcal{C}$ between the codebook $\mathcal{C}^T$ that stores class-related knowledge extracted from the labeled data, and then aggregates to refine coarse referring text, the similarity scores is calculated by
\begin{equation}
    W_{k}^\mathcal{C} = \mathrm{COS}(Q^{\mathrm{CLUE}}, \mathcal{V}_k),\ \mathcal{V}_k \in \mathcal{C}^T,
\end{equation}
where the $\mathrm{COS}(\cdot)$ is the consine similarity function.
Next, we use these scores $W^\mathcal{C}$ to perform a weighted aggregation of the vectors in the codebook, generating a clue vector,
\begin{equation}
    \mathcal{V}^{\mathrm{CLUE}} = \sum_{k}^{|\mathcal{C}^T|} W_{k}^\mathcal{C} \mathcal{V}_k.
\end{equation}

We use the class word $\mathrm{X}_{l,i}^{T_{cls}}$ extracted from the referring text $\mathrm{X}_{l,i}^{T}$ to supervise the clue vector $\mathcal{V}^{\mathrm{CLUE}}$,
\begin{equation}
    \mathcal{L}_{\mathrm{TFM}}^{\mathrm{CLUE}} = \mathcal{L}_{\mathrm{MSE}}(\mathcal{V}^{\mathrm{CLUE}}, \mathbf{CLIP}(\mathrm{X}_{l,i}^{T_{cls}})).
\end{equation}

We combine this clue vector $\mathcal{V}^{\mathrm{CLUE}}$ with the referring text features $\mathrm{F}_{l,i}^{T}$ to enhance the representation ability. Then, we use a hierarchical feature cross-attention mechanism for text-visual information fusion,
\begin{equation}
\begin{gathered}
    \mathrm{F}_{l,i}^{{T'}} = \mathrm{Concat}(\mathrm{F}_{l,i}^{T}, \mathcal{V}^{\mathrm{CLUE}}) \\
    \mathrm{F}_{l, i, k}^{I'} = \mathrm{CrossAttn}(\mathrm{F}_{l, i, k}^{I}, \mathrm{F}_{l,i}^{{T'}}),
\end{gathered}
\end{equation}
where $\mathrm{Concat}(\cdot)$ denotes the concatenation operation, and $\mathrm{CrossAttn}(\cdot)$ denotes the cross-attention function. During training and inference of unlabeled data in $\mathcal{D}_{\mathrm{R}}$, only the attention scores $\mathrm{A}^{\mathrm{CAMO}}_{l, i}$ are supervised by the pseudo-labels from the teacher model. The total loss function for the TFM module can be formulated as,
\begin{equation}
    \mathcal{L}_{\mathrm{TFM}} = \mathcal{L}_{\mathrm{TFM}}^{\mathrm{Attn}} + \lambda_t\mathcal{V}^{\mathrm{CLUE}},
    \label{eq:ADA loss}
\end{equation}
where $\lambda_t = 1$ for labeled data in $\mathcal{D}_{\mathrm{L}}$ and $\lambda_t = 0$ for unlabeled data in $\mathcal{D}_{\mathrm{R}}$. We use the class words $\mathrm{X}_{l,i}^{T_{cls}}$ extracted from the referring text annotation $\mathrm{X}_{l,i}^{T}$ of the labeled data in $\mathcal{D}_{\mathrm{L}}$ to initialize the codebook $\mathcal{C}^T$.

\subsection{Total Loss}
\label{sec:totloss}

The total loss function can be represented as,
\begin{equation}
    \mathcal{L}_{tot} = \mathcal{L}_{\mathrm{s}} + \lambda_{\mathrm{u}}\mathcal{L}_{\mathrm{u}},
\end{equation}
where $\mathcal{L}_{\mathrm{s}}$ denotes the supervised loss, $\mathcal{L}_{\mathrm{u}}$ denotes the unsupervised loss, and $\lambda_{\mathrm{u}}$ is the weight of unsupervised loss to balance the loss terms. The labeled loss function consists of three parts,
\begin{equation}
    \mathcal{L}_{\mathrm{s}} = \mathcal{L}_{\mathrm{Seg}}^{\mathrm{s}} + \mathcal{L}_{\mathrm{Aug}} + \mathcal{L}_{\mathrm{TFM}},
    \label{eq:supervise loss}
\end{equation}
where $\mathcal{L}_{\mathrm{Seg}}^{\mathrm{s}}$ denotes the supervised segmentation loss between the student predictions and groud-truth. The unlabeled loss function consists of two parts,
\begin{equation}
    \mathcal{L}_{\mathrm{u}} = \mathcal{L}_{\mathrm{Seg}}^{\mathrm{u}} + \mathcal{L}_{\mathrm{TFM}},
    \label{eq:unsupervise loss}
\end{equation}
where $\mathcal{L}_{\mathrm{Seg}}^{\mathrm{u}}$ denotes the semi-supervised loss between the student predictions and pseudo labels (produced by the teacher model). Following \cite{Zheng2024}, we use a combination of Binary Cross Entropy loss $\mathcal{L}_{\mathrm{BCE}}$, Intersection Over Union loss $\mathcal{L}_{\mathrm{IOU}}$, and Structure Similarity Index Measure loss $\mathcal{L}_{\mathrm{SSIM}}$ to build the supervised loss $\mathcal{L}_{\mathrm{Seg}}^{\mathrm{s}}$ and semi-supervised loss $\mathcal{L}_{\mathrm{Seg}}^{\mathrm{u}}$. The complete definition of these loss functions can be found in the appendix.

\section{Experiments}

\begin{figure*}
    \centering
    \includegraphics[width=\linewidth]{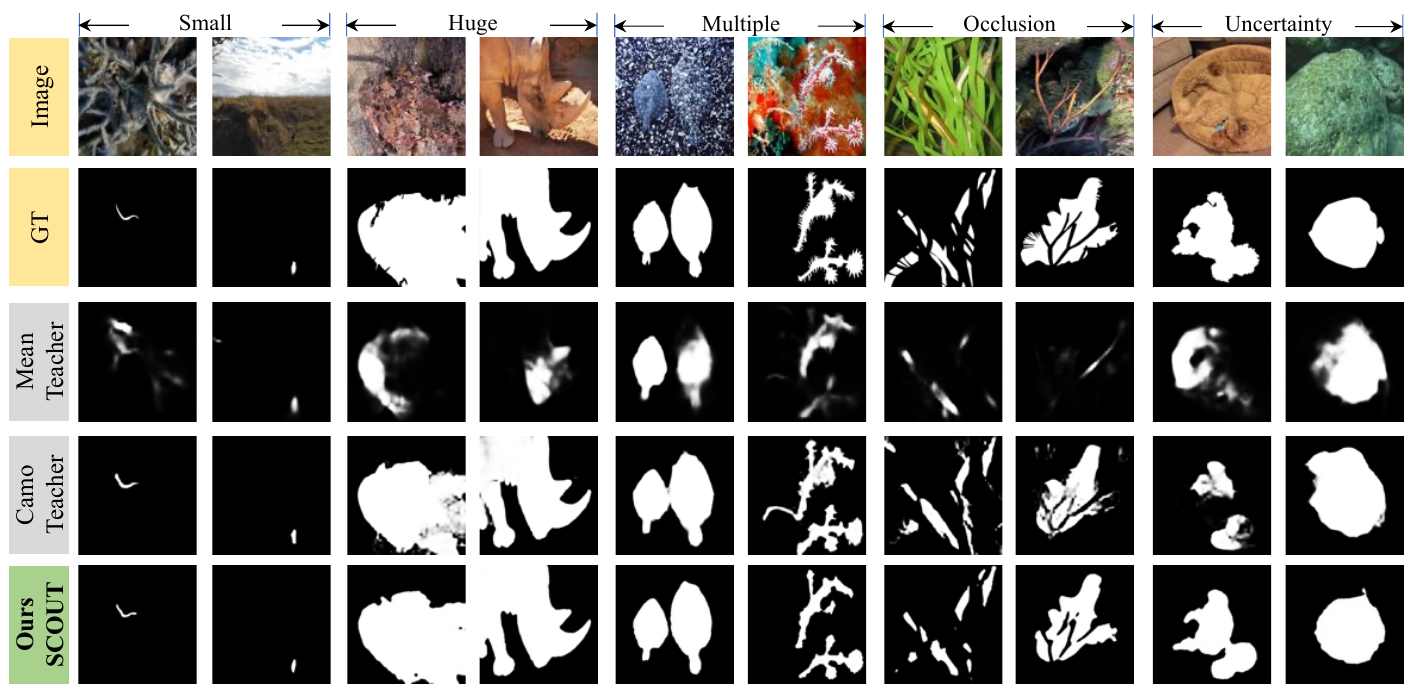}
    \caption{\textbf{Visual comparison of our method with other existing methods in challenging scenarios.} Our method has clearer and more precise segmentation boundaries and correctly recognizes depth-camoflaged objects.}
    \label{fig:main_visual}
\end{figure*}

\subsection{Experiment Settings}

\textbf{Training Set.} To compare with the existing works, following \cite{Luo2023,Fan2020b},
we use 1000 images from the CAMO trainset and 3040 images from the COD10K trainset as the training set for our experiments. During the training process, we follow the data partition ratios from previous semi-supervised COD results \cite{Lai2024}, training the model with 1\%, 5\%, and 10\% of labeled data. However, diverging from traditional semi-supervised segmentation approaches, we do not employ a random data sampling strategy. Instead, we utilize the proposed ADSA module to actively select the valuable data with annotating labels (\textit{i.e.} segmentation mask and referring text), while the remaining portion is treated as unlabeled data. For all unlabeled data, the referring text is fixed to a single sentence ``\textit{Camouflaged objects; hidden objects; concealed objects}".

\noindent\textbf{Testing Sets.} We test the model's performance on four mainstream COD benchmark testing sets, CHAMELEON with 76 test images, CAMO with 250 test images, COD10K with 2026 test images, and NC4K with 4121 test images. To comprehensively evaluate the model, we test its performance under two different settings: using fixed referring text (\textit{i.e.} ``\textit{Camouflaged objects; hidden objects; concealed objects}") and using precise image-level referring text during test.

\noindent\textbf{Evaluation Protocol.} For a fair and comprehensive evaluation, we employ the S-measure ($\mathcal{S}_m$) \cite{Fan2017}, mean and weighted F-measure ($\mathcal{F}_{\beta}^{m}$, $\mathcal{F}_{\beta}^{\omega}$) \cite{Margolin2014}, max and mean E-measure ($\mathcal{E}_{\epsilon}^{x}$, $\mathcal{E}_{\epsilon}^{m}$) \cite{Fan2018}, mean absolute error ($\mathcal{M}$) \cite{Perazzi2012}. 

\noindent\textbf{Implementation Details.} All images are resized to 640×640 for training and testing. We employ the ImageNet pre-trained Swin-base \cite{Liu2021} as our image encoder, use recently developed BiRefBlock \cite{Zheng2024} from High-Resolution Dichotomous Image Segmentation (HR-DIS) fields to build the decoder, and utilize CLIP-ViT-Large as our text encoder. The parameters of the CLIP text encoder are frozen during the training process, while all others are trainable. The batch size is set to 6 for each GPU during training, Adam is used as the optimizer, and the learning rate is initialized to 1e-4 and use multi-step decay strategy with 30 training epochs. All experiments are implemented with PyTorch 2.1 and a machine with Intel(R) Xeon(R) Silver 4214R CPU @ 2.40GHz, 256GiB RAM, and 8 NVIDIA Titan A800-80G GPUs. 

\subsection{Main Results}

\noindent\textbf{Qualitative Analysis.}
We show visualizations of a series of camouflaged object segmentation masks predicted by our method and related methods in some challenging scenarios. As shown in \cref{fig:main_visual}, we notice that our method achieves higher segmentation accuracy than existing semi-supervised COD approaches, with clearer edges and a more detailed representation of the camouflaged object. 

\noindent\textbf{Quantitative Analysis.}
We compare the proposed SCOUT's performance with existing semi-supervised models on four COD test datasets. 
Since these methods are not open-sourced and are difficult to reproduce, we only report the experimental results published in their original papers. 
As shown in \cref{tab:main_results}, SCOUT has already surpassed all previous methods across all metrics on all datasets when tested with fixed text (\textit{e.g.} ``\textit{Camouflaged objects; hidden objects; concealed objects; CLUE\_Token}''). Thanks to the assistance of the clue vector in TFM, SCOUT effectively improves fixed text, achieving results nearly identical to those inferences with precise texts.
Compared to the previous best results, our method achieves an improvement of 52.0\% in $\mathcal{M}$, 19.1\% in $\mathcal{S}_m$, 16.3\% in $\mathcal{E}_{\phi}^{m}$, and 29.1\% in $\mathcal{F}_{\beta}^{m}$, effectively demonstrating the high performance of the proposed SCOUT.

\subsection{Ablation Study}

\begin{table}
    \setlength{\belowcaptionskip}{0cm}   
    \renewcommand{\arraystretch}{1.1}
    \renewcommand{\tabcolsep}{3pt}
    \centering
    \resizebox{\linewidth}{!}{
    \begin{tabular}{ccc|cccccc}
    \toprule
    \multicolumn{3}{c|}{\textbf{Settings}}  & \multicolumn{6}{c}{\textbf{COD10K (2026)}} \\ \midrule
    \textbf{ADAS-ADA} & \textbf{ADAS-ADS} & \textbf{TFM} \metricsCOD{}  \\ \midrule
               &       &         &   .805      &      .685      &   .729    &   .856   &    .871  &   .038     \\
    \checkmark &            &            &    .824   &   .730   &    .765  &   .878   &    .900    &    .034  \\
               &            & \checkmark &  .822 & .720 & .772 & .869 & {.914} & .033 \\
    \checkmark &            & \checkmark &  .830 & .745 & .795 & .885 & .913& .029 \\
    \checkmark & \checkmark &            &  {.849} & {.763} & {.797} & {.905} & .913 & {.028} \\
    \checkmark & \checkmark & \checkmark & \textbf{.855} & \textbf{.768} & \textbf{.801} & \textbf{.913}  & \textbf{.924} & \textbf{.027}\\
     \bottomrule
    \end{tabular}}
    \caption{\textbf{Ablation study to evaluate the proposed modules.} We retrain our model with different settings under 5\% labeled data settings, and evaluation on the COD10K-Test sets.}
    \label{tab:abl_main}
\end{table}
\noindent\textbf{Module Ablations.}
To validate the effectiveness of the proposed module, we first perform ablations on the ADA and ADS components in the ADAS module and the TFM module. When the ADS component is not used, the labeled set is obtained through random sampling. When the TFM module is not used, the referring text does not participate in model training. We retrain the model with 5\% labeled data setting and test it on COD10K-Test set. The results are shown in \cref{tab:abl_main}, which indicates a significant performance drop when neither module is used. Compared to the baseline, our method achieves an improvement of 28.9\% in $\mathcal{M}$ and 6.21\% in $\mathcal{S}_m$.

\begin{table}
    \setlength{\belowcaptionskip}{0cm}   
    \renewcommand{\arraystretch}{1.1}
    \renewcommand{\tabcolsep}{3pt}
    \centering
    \resizebox{\linewidth}{!}{
    \begin{tabular}{cccc|cccccc}
    \toprule
    \multicolumn{4}{c|}{\textbf{Settings}}  & \multicolumn{6}{c}{\textbf{COD10K (2026)}} \\ \midrule
    \textbf{Rand-Color} & \textbf{Rand-Geo} & \textbf{Ada-Color} & \textbf{Ada-Geo} \metricsCOD{}  \\ \midrule
               &            &            &            &  .822 & .720 & .772 & .869 & .914 & .033     \\
    \checkmark &            &            &            &  .833 & .744 & .791 & .886 & .915 &.032 \\
               & \checkmark &            &            &  .837 & .737 & .766 & .899 & .907 &.031 \\
    \checkmark & \checkmark &            &            &  .846 & .757 & .796 & .901 & .918 &.030 \\
               &            & \checkmark &            &  .852 & {.768} & \textbf{.806} & .908 & \textbf{.927} &{.027} \\
               &            &            & \checkmark &  .841 & .750 & .787 & .901 & .913 &.029 \\
               &            & \checkmark & \checkmark &  \textbf{.855} & \textbf{.768} & .801 & \textbf{.913}  & \.924 & \textbf{.027} \\
    
     \bottomrule
    \end{tabular}}
    \caption{\textbf{Ablation study on different data augmentations.} ``Rand'' denotes the random augmentation, while ``Ada'' denotes the learnable augmentations. ``Color'' represents the color augment operation, and ``Geo'' represents the geometry augment operation. We retrain each model on different combinations of the augmenters under 5\% labeled data settings.}
    \label{tab:abl_adas}
\end{table}

\noindent\textbf{Effectiveness of Data Augment.}
We conduct comparative experiments between random augmentation and the trainable augmenters used in this paper. We retrain the model on different settings with 5\% labeled data and test on the COD10K-Test set. As shown in \cref{tab:abl_adas}, when using the learnable augmentation, we find that color augmentation has a significant effect. This is related to the complex colors typically found in camouflaged objects. 

\begin{table}
    \setlength{\belowcaptionskip}{0cm}   
    \renewcommand{\arraystretch}{1.1}
    \renewcommand{\tabcolsep}{6pt}
    \centering
    \resizebox{\linewidth}{!}{
    \begin{tabular}{c|cccccc}
    \toprule
    \multirow{2}{*}{\textbf{Settings}}  & \multicolumn{5}{c}{\textbf{COD10K (2026)}} \\ \cline{2-7}
     \metricsCOD{}  \\ \midrule
    Top-K Easy (Close to 1) &  .836 & .715 & .756 & .879 & .909 &.035 \\
    Top-K Hard (Close to 0) &  .845 & .747 & .770 & .912 & .920 &.028 \\
    Center-0.75&  .820 & .725 & .767& .880 & .910 &.030 \\
    Center-0.25&  .834 & .751 & .797 & .886 & .915 &.027 \\
    Center-0.5 &  \textbf{.855} & \textbf{.768} & \textbf{.801} & \textbf{.913} & \textbf{.924} &\textbf{.027} \\
    
     \bottomrule
    \end{tabular}}
    \caption{\textbf{Ablation study on different data selection strategies.} ``Top-K Easy/Hard'' denotes the selection of data by sampling in ascending/descending order. ``Center-$x$'' denotes the selection method where data is sampled symmetrically around a center, with uniform sampling both before and after the center $x$. We retrain the model under 5\% labeled data settings and test on the COD10K-Test set.}
    \label{tab:abl_ads_interval}
\end{table}

\begin{table}
    \setlength{\belowcaptionskip}{0cm}   
    \renewcommand{\arraystretch}{1.1}
    \renewcommand{\tabcolsep}{3pt}
    \centering
    \resizebox{\linewidth}{!}{
    \begin{tabular}{cc|cc|ccccccc}
    \toprule
    \multicolumn{2}{c|}{\textbf{Train}} & \multicolumn{2}{c|}{\textbf{Test}} & \multicolumn{6}{c}{\textbf{COD10K (2026)}} \\ \midrule
    \textbf{Fixed.} & \textbf{Precise.} & \textbf{Fixed.} & \textbf{Precise.} \metricsCOD{}  \\ \midrule
    \checkmark &            & \checkmark &            &   {.845} &   .765 & {.803} &  {.901} & .917 & {.027} \\
    \checkmark &            &            & \checkmark &  .843 & .761 &  \textbf{.806} & .899 & {.926} & .028 \\
               & \checkmark & \checkmark &        &      {.855} & {.768} & {.801} & {.913}  & {.924} &  {.027}\\
               & \checkmark &            & \checkmark &      \textbf{.859} &  \textbf{.776} & .804 & \textbf{.919} &  \textbf{.926} & \textbf{.026} \\
     \bottomrule
    \end{tabular}}
    \caption{\textbf{Ablation study to evaluate the different settings on referring text.} We retrain the model under 5\% labeled data setting with fixed and precise referring text for labeled data, then test on COD10K-Test dataset with fixed and precise referring text.}
    \label{tab:abl_reftext}
\end{table}

\noindent\textbf{Effectiveness of Data Scoring.}
We conduct an ablation experiment on the data sampling strategy used in the ADS component in the ADAS module. Specifically, we resample 5\% of the data by using Top-K easy, Top-K hard, and different score selection centers, and train the model. We then test it on the COD10K-Test set, and the results are shown in \cref{tab:abl_ads_interval}. When only easy and hard samples are used, the model struggles to achieve optimal performance. However, when selecting data around 0.5, the selected data not only suits the model's learning but also balances difficulty, resulting in the best performance.

\noindent\textbf{Effectiveness of Referring Text.}
We conduct further exploration into the role of referring text. We compare the performance of using precise and fixed referring text during both training and testing on the unlabeled set. The results are shown in \cref{tab:abl_reftext}. We find that precise referring text indeed helps the model learn camouflage-related knowledge more effectively, and the TFM is able to enhance fixed text, achieving performance comparable to that of precise text.

\section{Conclusions}

In this paper, we address the shortcomings of existing semi-supervised COD methods, which fail to adaptively select and utilize high-quality data, resulting in poor performance. We propose an innovative semi-supervised COD model SCOUT. Specifically, we propose the ADAS module to avoid meaningless data annotations by selecting valuable data through an adversarial augment and sampling strategy. Additionally, we propose the TFM to fully utilize the referring text by combining camouflage-related knowledge and text-visual interaction. Furthermore, we propose a RefTextCOD dataset, which contains a large number of image-level referring text annotations. Extensive experiments demonstrate the effectiveness of the proposed framework and modules.

\section*{Ethical Statement}
There are no ethical issues. 

\section*{Acknowledgements}
This work was supported by the National Science Fund for Distinguished Young Scholars (No.62025603), the National Natural Science Foundation of China (No. U21B2037, No. U22B2051, No. U23A20383, No. 62176222, No. 62176223, No. 62176226, No. 62072386, No. 62072387, No. 62072389, No. 62002305 and No. 62272401), and the Natural Science Foundation of Fujian Province of China (No. 2021J06003, No. 2022J06001).

\bibliographystyle{named}
\bibliography{ijcai25}

\end{document}